\documentclass[10pt, a4paper]{article}
\usepackage{lineno}
\usepackage[final]{lrec2026}
\usepackage{times}
\usepackage{latexsym}
\usepackage{comment}
\usepackage{tikz}
\usepackage{multirow}
\usepackage[edges]{forest}
\usepackage{pifont} 
\usepackage[table]{xcolor}
\usepackage{soul}

\newif\ifreviewhighlight
\reviewhighlighttrue

\sethlcolor{white!0}

\usepackage[utf8]{inputenc}

\usepackage{microtype}

\usepackage{inconsolata}

\usepackage{graphicx}
\usepackage{stfloats} 
\usepackage{float}
\usepackage{tikz}
\usepackage[edges]{forest}
\usepackage{pgfplots}
\usepackage{bm}
\usepackage{scalefnt}

\usepackage{hyperref}

\usepackage[framemethod=tikz]{mdframed}
\usepackage{subcaption}
\definecolor{lgreen}{rgb}{0.89,0.94,0.85}
\definecolor{lred}{rgb}{0.98, 0.90, 0.84}
\definecolor{lyellow}{rgb}{1.00, 0.95, 0.80}
\definecolor{lblue}{rgb}{0.85, 0.89, 0.95}
\definecolor{hidden-draw}{RGB}{20,68,106}
\definecolor{hidden-pink}{RGB}{255,245,247}

\usepackage{titlesec}
\titlespacing*{\section}
  {0pt}   
  {0.5em} 
  {0.25em} 

\titlespacing*{\subsection}
  {0pt}
  {0.4em}
  {0.1em}

\title{Automatic Analysis of Collaboration Through Human Conversational Data Resources: A Review}

\name{Yi Yu\textsuperscript{1,2}, Maria Boritchev\textsuperscript{2}, Chloé Clavel\textsuperscript{1,2}} 

\address{\textsuperscript{1}INRIA Paris, ALMAnaCH, \textsuperscript{2}LTCI, Télécom Paris, Institut Polytechnique de Paris, France \\
         48, rue Barrault 75013 Paris, 19 place Marguerite Perey, 91120 Palaiseau \\
         \{yi.yu, chloe.clavel\}@inria.fr, maria.boritchev@telecom-paris.fr}

\abstract{
Collaboration is a task-oriented, high-level human behavior. In most cases, conversation serves as the primary medium for information exchange and coordination, making conversational data a valuable resource for the automatic analysis of collaborative processes. In this paper, we focus on verbal aspects of collaboration and conduct a review of collaboration analysis using task-oriented conversation resources, encompassing related theories, coding schemes, tasks, and modeling approaches. We aim to address the question of how to utilize task-oriented human-human conversational data for collaboration analysis. We hope our review will serve as a practical resource and illuminate unexplored areas for future collaboration analysis.
 \\ \newline }

\begin{document}
\maketitleabstract

\section{Introduction}
\label{sec:introduction}

Collaboration analysis (CollA) seeks to use computational methods to model how people coordinate, think, and learn in shared tasks in order to gain insights that improve both collaboration processes and outcomes \cite{martinez2021you}. As collaboration is a fundamental human behavior, CollA has broad applications, including education \cite{jaques2023using}, management \cite{casey2009sticking}, interface design \cite{prati2021use}, and AI agent development \cite{enayet2023proactive,zhang-etal-2024-exploring}.

\begin{figure}[ht]
    \centering
    \includegraphics[width=\columnwidth]{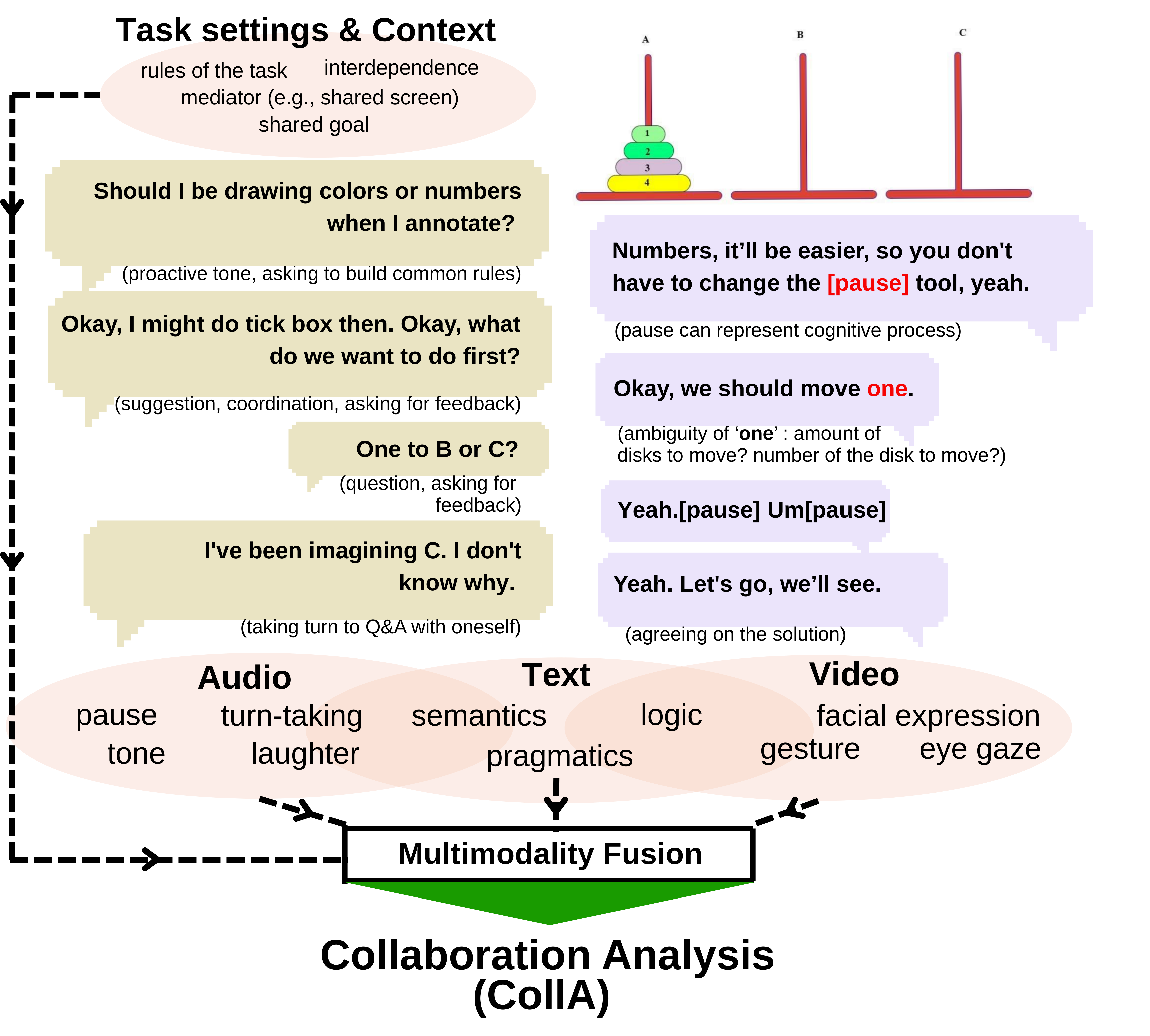}
    \caption{Two players playing the Hanoi tower game and conversing with each other to solve the puzzle. Their conversation provides many useful elements for CollA.}
    \label{fig:1}
\end{figure}

Computational methods for CollA require data to understand the phenomena in play. Human conversational resources are irreplaceable for two main reasons. 
{First}, conversations are instances of joint action~\cite{clark1996using}, and human conversational data is a sequential record of multimodal communicative behavior that reflects both individual contributions and interpersonal dynamics.
Additionally, linguistic research on human interpersonal phenomena supplies features for CollA, such as referring expressions \cite{heeman1995collaborating,clark1986referring} and multilevel entrainment \cite{lubold2014acoustic}, as discussed in Section \ref{sec:features}.
We illustrate in Figure \ref{fig:1} how informative human task-oriented conversation can be for CollA. 
{Second}, although we have relatively mature metrics for evaluating task-related dimensions of CollA in task-oriented conversations (\textit{e.g.}, task decomposition, task completion, etc. \citet{guan2025evaluating}), interpersonal dynamics, which directly affect the collaboration process and quality, remain largely unexplored.
For collaborative learning scenarios, collaboration itself can be a way of learning. Collaboration directly fosters both the internalization and the sharing of knowledge, which provide new dimensions for CollA. In addition, the quality of interpersonal dynamics can significantly influence learning outcomes \cite{yang2023historical}.
The variety of interaction patterns in human task-oriented dialogue also makes it possible to investigate how people come to know each other (\textit{e.g.}, through language use, personality traits, or educational background) during collaboration \cite{guo2025user}. 
CollA using human conversational data provides valuable insights for designing human-machine collaboration systems that balance individual benefits with overall task performance of the group, and provide a better understanding of humanity.

Several surveys have covered related but distinct topics: 
\citet{praharaj2021literature} and \citet{schurmann2024conceptualization} examine CollA multimodal features and existing collaboration measurements, but focus on the educational context. 
\citet{zou2025survey} centers on LLM-based
human-agent system building and examines how human feedback and control contribute to performance improvement. They review works on human-LLM systems and the conversation corpora for these systems, focusing on tasks where human feedback can be crucial for defining and assessing evaluation metrics. 
\citet{vaccaro2024combinations} surveys recent work comparing the performance of humans working alone, humans collaborating with machines, and machines operating independently, and highlights the challenges involved in integrating human intelligence with computational systems. While this study investigates the conditions under which human–machine collaboration surpasses human-only or machine-only performance, our work instead reviews computational models of multimodal discourse in collaborative contexts.

A holistic review of human task-oriented conversation resources and their usage for CollA remains unexplored. 
By task-oriented conversation, we mean conversation directed with clear intention toward the completion of  tasks \citet{grosz1986attention}.   
For collaboration, building on the definitions provided by \citet{wood1991toward,randrup2016philosophy}, we conceptualize collaboration as an interactive process comprising four core elements: \textbf{a shared goal}, \textbf{a shared understanding of the task} (including rules, norms, and structures), \textbf{positive interdependency}, and \textbf{joint individual commitments} reflected in participants' actions and decisions. 
Then, we apply this definition to select task-oriented corpora from peer-reviewed published papers that provide collaboration annotations and evaluations, focusing particularly on settings that create positive interdependencies among participants (\textit{e.g.}, tasks that cannot be completed by a single participant under the defined settings). The step-by-step procedure and criteria for paper selection are explained in Appendix A, Figure \ref{fig:data_select}. We conduct the review by systematically analyzing the coding schemes for capturing collaboration in conversation, extracting salient multimodal features, and examining recent collaboration modeling approaches applied to the selected corpora.

After discussing coding schemes in Section \ref{sec:coding}, we present the criteria used to select CollA corpora and review their task settings in Section \ref{sec:corpora}. We then examine CollA studies based on at least one of the selected corpora, discussing salient features and modeling approaches in Sections \ref{sec:features} and \ref{sec:models}. We conclude by discussing recent advances and future directions of CollA in Section \ref{sec:concl}.

 
\section{Coding Schemes for CollA}
\label{sec:coding}

Coding schemes serve as a lens to showcase important elements for CollA \cite{chen2020coding} and are used to annotate collaboration in conversations for computational collaboration model building. 
This section reviews coding schemes employed in the corpora discussed in Section \ref{sec:corpora}, highlighting how different coding approaches capture both individual and group (including dyad) aspects of collaboration. We also examine applied questionnaires, given their flexibility (\textit{e.g.}, self-reports, annotation instruments, external evaluation) and wide application in CollA. 

The full details of all the employed coding schemes are given in Table \ref{tab:table_coding_v2}. We discuss the main categories of coding schemes and questionnaires, their theoretical foundations, and how they capture different aspects of collaboration to illuminate current trends and challenges.

\subsection{Individual Perspective}
From the individual perspective,  applied coding schemes and questionnaires
focus on the single-participant collaborative aspect, such as \textit{individual collaborative/cooperative behaviors}, \textit{engagement}, and individual variables, such as \textit{collective orientation} that refers to the propensity to work in a collective manner in a team setting \cite{driskell2010collective}. 
They are applied to individual audio and visual data to model participants' collaborative behaviors and their impacts on the collaboration process.

Using \textbf{linguistic} aspects, 
\citet{cavicchio2012rovereto} use a \textit{cooperation} coding scheme in \hyperlink{coding_row:rovereto}{Rovereto} corpus, developped by \citet{davies1997empirical,davies2007grice}. This scheme analyzes individual collaborativeness using Grice's cooperative principle \cite{grice1975logic} for conversation analysis. It is evaluative, \textit{i.e.}, not only used to label what the speakers do but also to assess it in terms of appropriateness, which can be subjective and hard to agree on for the annotators. 
Findings in classroom discourse research reveal the important role of individual argument in the collaborative learning process \cite{engle2002guiding}. Based on that, \citet{olshefski2020discussion} develop a coding scheme to capture the function of collaborative argument moves for students' discussion in the \hyperlink{coding_row:discuss}{Discussion Tracker} corpus to explore the collaboration dimension of argument in large-group collaborative learning tasks.

Using \textbf{behavior} studies, \citet{richey2016sri} applied a modified version of the collaborative behavior coding scheme from \citet{johnson2013cooperation} 
to the \hyperlink{coding_row:sri}{SRI corpus}. These coding schemes are rooted in social interdependence theory \cite{johnson2009educational} and Vygotsky's cognitive developmental theory \cite{tudge2014peer}, which highlight the interactive aspects of individual behavior and collaborative indicators of learning. \citet{reverdy2022roomreader} adapts a multifacet classroom engagement behavior coding scheme \cite{goldberg2021attentive} for online interaction in the \hyperlink{coding_row:room}{RoomReader} corpus, where head and hand position and eye gaze/focus play an important role in annotating individual engagement levels. \citet{peechatt2024multicollab} design a coding scheme for annotating the frustration level in \hyperlink{coding_row:collab}{MULTICOLLAB} to predict critical moments in collaboration process.

From a \textbf{human factors} perspective, \citet{litman2016teams} collect collective orientation ("the propensity to work in a collective manner", \citet{driskell2010collective}) from each participant via a self-assessment questionnaire in the \hyperlink{coding_row:teams}{Teams} corpus, enabling further collaboration analysis alongside individual variables.
\setlength{\tabcolsep}{2pt}

 \begin{table*}[ht]
  \tiny
    \centering
    \begin{tabularx}{\textwidth}{|p{0.3cm}|p{3cm}|p{3.2cm}|p{6.6cm}|p{2cm}|}
     \cline{2-5}
    \multicolumn{1}{c|}{} &\textbf{Corpus, Description} &\textbf{Scheme/Questionnaire} &\textbf{Details} &\textbf{Reference}\\ \hline

    \multirow{6}{0.3cm}{\rotatebox{90}{\parbox{2.3cm}{\centering \textbf{Individual}}}} & \hypertarget{coding_row:sri}{\textbf{SRI} speech-based collaborative learning corpus} &  I Code (individual collaboration indicators) & regulative/logistical, interaction, and cognitive indicators of teamwork behavior&\citet{richey2016sri}\\ \cline{2-5}

    & \hypertarget{coding_row:misc}{\textbf{MISC} information-seeking conversations}  & adapted User Engagement Scale (UES) & scaled ratings of partner's collaborativeness & \citetlanguageresource{mcduff2017multimodal}
    \\ \cline{2-5}
    
    &\hypertarget{coding_row:room}{\textbf{RoomReader} multimodal, multiparty conversational interactions corpus} & online engagement &  continuous scaled engagement in groups based on collaborators' behaviors and perceived intentions  &\citet{reverdy2022roomreader}\\ \cline{2-5}
    
    &\hypertarget{coding_row:discuss}{\textbf{Discussion Tracker} multiparty discussions}  & collaborative argumentation functions & classification of arguments as: new ideas, agreements, extensions, probes/challenge & \citetlanguageresource{olshefski2020discussion} \\ \cline{2-5}
    
    &\hypertarget{coding_row:rovereto}{\textbf{Rovereto} emotion and cooperation corpus} & cooperative dialogue effort &  evaluation of each turn along three dimensions: knowledge sharing, non-cooperative behavior, and cooperation level (scaled) &\citetlanguageresource{cavicchio2012rovereto}\\ \cline{2-5}
    
    &\hypertarget{coding_row:collab}{\textbf{MULTICOLLAB} multimodal dialogues} & extreme emotion (frustration) in collaboration &  participant self-assessed frustration level &\citet{peechatt2024multicollab}\\ \cline{2-5}
    
    &\hypertarget{coding_row:teams}{\textbf{Teams} multiparty dialogues for entrainment} & collective orientation & participant self-assessed preference for teamwork &\citet{litman2016teams}
    \\ \cline{1-5}

    \multirow{8}{0.3cm}{\rotatebox{90}{\parbox{2.5cm}{\centering \textbf{Dyad \& Group}}}}& \hypertarget{coding_row:sri_group}{\textbf{SRI} 
    } & Q Code (team collaboration quality; triads) & team-level quality states (\textit{e.g.}, good collaboration, follow-the-leader), based on number of engaged participants &\citet{richey2016sri}
    \\ \cline{2-5}
    & \hypertarget{coding_row:misc_group}{\textbf{MISC}}  & adapted User Engagement Scale (UES) & scaled ratings of collaboration process & \citet{mcduff2017multimodal}\\ \cline{2-5}
    
    & \hypertarget{coding_row:teams_group}{\textbf{Teams}}  & team cohesion, satisfaction, potency/efficacy &  between and post-game questionnaires elicit perceptions of teams processes&\citet{litman2016teams}\\ \cline{2-5}
    
    &\hypertarget{coding_row:game}{\textbf{GAME-ON} group analysis of multimodal expression of cohesion corpus}  & modified Group Environment Questionnaire (GEQ) for group cohesion & highlights of instrumental function (social vs task facets); affective function optional depending on study &\citet{maman2020game}\\ \cline{2-5}
    
    &\hypertarget{coding_row:gap}{\textbf{GAP} group affect and performance corpus} & teamwork experience (self-report) & ratings of teamwork performance (time management, efficiency, overall work quality) & \citetlanguageresource{braleyGroupAffectPerformance2018} \\ \cline{2-5}
    
    &\hypertarget{coding_row:ami}{\textbf{AMI} augmented multiparty interaction corpus, \textbf{PCC} patient consultation corpus} & \multirow{2}{3cm}{group cohesion} & \multirow{2}{8cm}{ratings of task cohesion, social cohesion, and leadership} &\citet{hung2010estimating}
    \\ \cline{5-5}
    &&  & &  \citet{kantharaju2021social} \\ \cline{2-5}
    
    &\hypertarget{coding_row:simo}{\textbf{MULTISIMO} multimodal group interaction corpus}  & collaboration quality (overall)  & scaled rating of overall collaboration quality & \citetlanguageresource
    {koutsombogera2018modeling}  \\ \cline{2-5}
    &\hypertarget{coding_row:photo}{\textbf{PhotoBook} visually-grounded dialogues}  & collaboration performance & scaled ratings of overall collaboration performance and perceived mutual understanding &\citet{haberPhotoBookDatasetBuilding2019}  \\ \cline{1-5}
    
    \end{tabularx}
    \caption{Applied coding schemes that capture different aspects of collaboration in conversational data.}
    \label{tab:table_coding_v2}
\end{table*}

\subsection{Dyad and Group Perspective}
The dyad and group interaction dynamics of the collaboration process have attracted lots of attention from the research community. These coding schemes or questionnaires focus on the interpersonal dynamics and group behaviors, such as 
\textit{group cohesion}, 
self-assessed \textit{collaboration performance}, 
and perceived \textit{collaboration quality}. 

For collaboration quality evaluation, \citet{haberPhotoBookDatasetBuilding2019} use a questionnaire in the dyadic conversation corpus \hyperlink{coding_row:photo}{PhotoBook} to collect self-assessed, scaled overall collaboration performance, while \citet{mcduff2017multimodal} select items from the \texttt{User Engagement Scale} questionnaire \cite{o2010development} in \hyperlink{coding_row:misc_group}{MISC}, for collaboration process evaluation. For three-person groups, \citet{richey2016sri} develop a coding scheme, \texttt{Q codes} (as shown in Table.~4 of \citet{richey2016sri}), in which the 
collaboration quality is defined as ``nb of the group members' actively contribute to the task completion'' with a special focus on balanced involvement of each member.

Group cohesion is a group phenomenon defined as ``the group members' inclinations to forge social bonds'' \citet{casey2009sticking}, which impacts the collaboration process \cite{hung2010estimating, kantharaju2021social}.
\citet{hung2010estimating} study both the social and task aspects of cohesion in a computational way. Their 27-item questionnaire for perceived group cohesion is based on group research \cite{carron2000cohesion} and psychology literature \cite{siebold1999evolution}, and it has been applied to the \hyperlink{coding_row:ami}{AMI} corpus \cite{kraaij2005ami} and the \hyperlink{coding_row:ami}{PCC} corpus \cite{kantharaju2021social}. 

\citet{severt2015function} provides another coding scheme for a cohesion study that includes functional and structural dimensions of cohesion. This scheme has two psychological functions: \textit{affective} and \textit{instrumental}, but only the last one, with \textit{social} and \textit{task} facets, has been used in the \hyperlink{coding_row:game}{GAME-ON} corpus \cite{maman2020game}, to align with the dominant approach~\cite{braun2020exploring}. 

\subsection{Main Trends for CollA Coding Schemes}
There is no universal coding scheme for CollA. For segment level, earlier studies tend to provide annotations for individual-level collaborativeness, while recent coding schemes are mostly applied to study group-level collaborative phenomena. 
One possible explanation could be that a group is not simply the sum of its dyads. Group-level collaboration involves emergent dynamics that individual or dyadic models cannot capture. 
However, group-level emergent collaborative interaction is context-dependent and can be hard to capture with individual-level cues. Human manual annotation remains the most reliable approach for these studies, given that large language models (LLMs) are increasingly involved in recent annotation processes \cite{wang2024human}.

We observe that task-level annotation has been widely chosen to evaluate both individual-level \cite{mcduff2017multimodal,haberPhotoBookDatasetBuilding2019} and group-level \cite{koutsombogera2018modeling, braleyGroupAffectPerformance2018,litman2016teams} collaboration. 
Most annotations are based on task-adapted questionnaires, such as perceived collaboration quality using \texttt{User Engagement Scale} \cite{o2010development} in \hyperlink{coding_row:misc}{MISC} and self-assessed overall collaboration quality and satisfaction in \hyperlink{coding_row:simo}{MULTISIMO}, \hyperlink{coding_row:gap}{GAP}, and \hyperlink{coding_row:teams_group}{Teams}. Both external and self-assessed annotations are valuable for CollA, from modeling collaborative conversation to detecting individual variables for dialogue system adaptation.
However, task-level granularity can be insufficient for analyzing emergent phenomena.

\section{Task-Oriented Corpora for CollA}
\label{sec:corpora}

In this section, we examine \textbf{open-source}, \textbf{human-human}, \textbf{task-oriented} conversation corpora created \textbf{in the last 20 years (from 2005)} that \textbf{require collaboration} in their task-settings and have \textbf{direct annotations of collaboration} (\textit{e.g.}, collaboration quality/skills, group cohesion, conflicts, etc) to understand the recent trends in task setting design for CollA. Corpora with only measurable collaboration task results will not be included here, \textit{e.g.} ELEA \cite{sanchez2011audio}, since task results alone cannot reflect collaboration quality.
\setlength{\tabcolsep}{2pt}
\begin{table*}[ht]
  \tiny
    \centering
    \begin{tabularx}{\textwidth}{p{1.5cm}p{0.5cm}p{0.7cm}p{0.7cm}p{0.5cm}p{0.5cm}p{0.5cm}p{0.5cm}p{0.7cm}|p{2.5cm}p{1.2cm}p{1.2cm}p{1.2cm}|p{1.5cm}}
     \hline
     \multirow{2}{1.5cm}{\textbf{Corpus}}
     &\multirow{2}{0.5cm}{\textbf{Lang.}}
     &\multirow{2}{0.7cm}{\textbf{Hours}} 
     &\multirow{2}{0.5cm}{\textbf{Task Type}} 
     &\multirow{2}{0.5cm}{\textbf{Group Size}}
     &\multirow{2}{0.5cm}{\textbf{Audio}}
     &\multirow{2}{0.5cm}{\textbf{Video}}
     &\multirow{2}{0.5cm}{\textbf{Transc.}}
     &\multirow{2}{0.5cm}{\textbf{Sensors}} 
     &\multicolumn{4}{c|}{\textcolor{blue}{\textbf{Collaboration Annotation}}}&\multirow{2}{1.5cm}{\textbf{Dataset Link}}
     \\ \cline{10-13}
     
     &&&&&&&&&\textcolor{blue}{\textbf{Object}}&\textcolor{blue}{\textbf{Obj. Level}}&\textcolor{blue}{\textbf{Annotator}} & \textcolor{blue}{\textbf{Granularity}} & \\ \hline
     
     \hypertarget{data_row:ami}{\textbf{AMI 2005}} &EN& 100 & 
      meeting & 4 & \ding{51}&\ding{51}&\ding{51}&\ding{55}&cohesion& group&external&segment&\href{https://groups.inf.ed.ac.uk/ami/download/}{Dataset link} \\ \hline

     \textbf{Rovereto 2012} & IT & 4.67 & game & 4 & \ding{51} & \ding{51} & \ding{51}&\ding{51} & cooperativeness& individual &external&segment& \href{https://www.protocols.io/de/view/recc-corpus-stimulus-set-and-protocol-j8nlkwzq5l5r/v1}{Dataset on request} \\ \hline
     
     \hypertarget{data_row:sri}{\textbf{SRI 2016}} & EN & 26.6 & education &3& \ding{51}&\ding{55}&\ding{51}&\ding{55}& I Code, Q Code& both& external &segment & \href{https://catalog.ldc.upenn.edu/LDC2019S01}{Dataset link}\\ \hline

     \ding{80}\hypertarget{data_row:teams}{\textbf{Teams 2016}} &EN &47&game&3-4&\ding{51}&\ding{51}&\ding{51}& \ding{55}& participant's collective orientation; group cohesion, satisfaction, potency& both& self&task & \href{https://sites.google.com/site/teamentrainmentstudy/corpus}{Dataset link}\\ \hline

     \ding{96}\hypertarget{data_row:misc}{\textbf{MISC 2017}} &EN &42& game&2& \ding{51} & \ding{51}&\ding{51}auto&\ding{55}&perceived collaboration (help/ understanding/communication)&individual&external&task& \href{https://www.microsoft.com/en-us/download/details.aspx?id=55594}{Dataset link}\\ \hline
     
     \textbf{MULTISIMO 2018} & EN &4 & game & 3  & \ding{51}&\ding{51}&\ding{51}&\ding{55} & collaboration quality & group & external& task& \href{https://multisimo.eu/datasets.html}{Dataset link}\\ \hline
     
     \hypertarget{data_row:gap}{\textbf{GAP 2018}} & EN & 4  & meeting & 2,3,4 & \ding{51}&\ding{55}&\ding{51}&\ding{55} & teamwork experience & group & self & task& \href{https://github.com/gmfraser/gap-corpus}{Dataset link}\\ \hline

    
     \hypertarget{data_row:photo}{\textbf{PhotoBook 2019}} &EN&-&game &2&\ding{55} &\ding{55}&\ding{51}&\ding{55}& collaboration performance& group & self & task& \href{https://dmg-photobook.github.io}{Dataset link} \\ \hline 

     \hypertarget{data_row:game}{\textbf{GAME-ON 2020}} &IT&11& game & 3 & \ding{51}& \ding{51}&\ding{55}&\ding{51}& cohesion&group&self&task & \href{https://grace.wp.imt.fr/the-game-on-dataset/}{Dataset link}\\ \hline 

    \hypertarget{data_row:dt}{\textbf{Discussion Tracker 2020}} &EN& - & education&15 & \ding{55}&\ding{55}&\ding{51}&\ding{55}& collaborative argumentation& individual& external&segment & \href{https://discussiontracker.cs.pitt.edu/\#top}{Dataset link}\\ \hline

     \hypertarget{data_row:pcc}{\textbf{PCC 2021}} &EN&2& other & 3,4 & \ding{51}&\ding{51}&\ding{55}&\ding{55} &cohesion& group & external &segment& \href{https://pcc.arg.tech/}{Dataset on request}\\ \hline


     \hypertarget{data_row:room}{\textbf{RoomReader 2022}}&EN &38& education &4,5 & \ding{51}&\ding{51}&\ding{51} &\ding{55} &engagement, cohesion & both & both&segment &\href{https://www.sigmedia.tv/datasets/room_reader/}{Dataset link}  \\ \hline


      \ding{80}\hypertarget{data_row:colla}{\textbf{MULTICOLLAB 2023}} & EN &3& other &2 & \ding{51}&\ding{51}&\ding{51}&\ding{51} & extreme emotion (frustration) & individual & self & both & \href{https://github.com/mp6510/MULTICOLLAB}{Dataset will go public}\\ \hline
    
      \multicolumn{14}{l}{\ding{96} The transcripts in MISC are auto-generated without mentioning any manual verification process.} \\ \hline
      \multicolumn{14}{l}{\ding{80}MULTICOLLAB is currently not public but is to be made available in the future. The video, transcription, and questionnaire data of Teams will be available in a future release.} \\ \hline

    \end{tabularx}
\caption{Overview of collaboration corpora arranged chronologically, to highlight the evolution of research focus in CollA. Comparative 
assessment across dimensions: language, recording size, task type, group size, multimodal data availability, and collaboration annotation (annotation object, object level, annotator type, and temporal granularity). 
}
    \label{tab:table_corpora}
\end{table*}

The task settings are vital for building CollA corpora, as they determine the level of interdependence among participants during the collaboration process. We first discuss CollA task settings, including group size and how they promote collaboration between participants, and provide details on their collaboration annotations in Table \ref{tab:table_corpora}. We then compare different choices and synthesize the main trends.

\vspace{-0.5em}

\vspace{-0.4cm}
\paragraph{Game}
Game scenarios constitute the dominant task setting category in our selection of corpora. We categorize a corpus task as a game when the keyword ``game'' is used to describe the corpus' scenario. 

Social game settings have been intensively used for group cohesion analysis. The \hyperlink{data_row:teams}{Teams} corpus \citeplanguageresource{litman2016teams} is built on a role-playing social game for CollA. A group of 3 or 4 players, each with a different adventurer role, must discuss strategies to collect enough treasures to complete the game. 
The \hyperlink{data_row:game}{GAME-ON} corpus \citeplanguageresource{maman2020game} is based on a multitask social game in which 3 players must cooperatively discover clues and solve several puzzles within a limited time frame. Perceived cohesion is evaluated through self-assessment after each puzzle. They argue that cohesion can take considerable time to emerge in groups of strangers, so they recruit only real-world friends to play the game. 

Game settings can be easily adapted to a particular aspect of CollA. The \hyperlink{data_row:misc}{MISC} corpus \citeplanguageresource{mcduff2017multimodal} 
employs a role-play, information-seeking setting, assigning the ``seeker'' role to one participant in each pair. This setting helps understand human interests in this collaborative information-seeking process to design and evaluate human-machine
interfaces.
The \hyperlink{data_row:photo}{PhotoBook} corpus \citeplanguageresource{haberPhotoBookDatasetBuilding2019} uses a remote multi-round image identification game setting.
Each group of two players has access to different sets of images, and participants must mark each image as either common or different by discussing it with their partner. This setting enables referring analysis\footnote{``John didn’t come to class because \textbf{he} was sick.'' Referring analysis studies what the word ``he'' refers to, tracking who or what is being talked about, a key part of collaborative conversation.} in a collaborative task.

\vspace{-0.5em}
\paragraph{Education}
Education is a domain where collaboration has been extensively studied in the context of collaborative learning.
Due to privacy concerns, a significant number of corpora under this task setting are not public
\cite{bertrand2022using_010,lama2021deep_011,olsen2020temporal_014,Spikol2017-ok_018,salinas2021can}. Recent corpora in collaborative learning settings often involve large-group interactions, making the capture of multimodal data more difficult. 

Among the accessible resources, the \hyperlink{data_row:dt}{Discussion Tracker} corpus \cite{olshefski2020discussion} captures classroom teacher-student interactions, focusing on collaborative argumentation in literature discussions. The \hyperlink{data_row:sri}{SRI} corpus \citeplanguageresource{richey2016sri} exemplifies collaborative problem-solving with triadic groups solving math problems, capturing both social and cognitive dimensions mainly from audio. 

The only corpus we found with video data and collaborative learning task setting is the \hyperlink{data_row:room}{RoomReader} corpus \citeplanguageresource{reverdy2022roomreader}. It is based on online computer-supported student-tutor conversations and can be used to analyze engagement in collaboration and conversational dynamics.

\vspace{-0.4cm}

\paragraph{Meetings and Others}
Scenarios adapted from real-world tasks are frequently used for CollA.

A part of the \hyperlink{data_row:ami}{AMI} corpus \citeplanguageresource{kraaij2005ami} elicits collaboration using a role-playing functional meeting within a 4-person design team for new product protocol development. 
The \hyperlink{data_row:gap}{GAP} corpus \citeplanguageresource{braleyGroupAffectPerformance2018} also applies a meeting setting for a small group to make a decision on the rank of the most important items for a plane crash. 
The \hyperlink{data_row:pcc}{PCC} corpus \citeplanguageresource{kantharaju2021social} simulates health consultations between patients and healthcare professionals by equipping professionals with detailed background information to study cohesion. 

The \hyperlink{data_row:colla}{MULTICOLLAB} corpus \citeplanguageresource{peechatt2024multicollab} adopts a role-playing setting, \textit{i.e.}, instructor and builder, for a block building task in which some builders are instructed to deliberately disobey to stimulate critical moments in the collaboration process with strong interdependency. Their task settings yield half of their data from non-collaborative builders.

\vspace{-0.5em}
\paragraph{Main Trends of CollA Task Settings}
Common task settings for CollA include role-play and information asymmetry, which promote interdependence among participants and foster the observation of active joint contributions during the collaboration process. 
Co-locating collaboration corpora with video recordings, especially with both individual- and room-level recordings, is interesting for studying interpersonal and group dynamic aspects of CollA \cite{kraaij2005ami, litman2016teams, koutsombogera2018modeling, maman2020game} and have attracted more attention in recent CollA. 

Remote meetings can also provide a view of group synchrony (\textit{e.g.}, \hyperlink{data_row:room}{RoomReader} \cite{reverdy2022roomreader}), while it is recognized that remote settings may inhibit natural interaction \cite{poel2008meeting}.
We also observe a trend toward analyzing group collaboration phenomena using segment-level annotations, rather than individual collaborativeness as in earlier CollA studies.

\section{CollA Features}
\label{sec:features}
This section synthesizes experimentally supported features extracted from text, audio, video, sensor signals\footnote{We find the following sensor signals applied for CollA: electrocardiogram, electrodermal signals, galvanic skin response, photoplethysmography, and body motion.}, and cross-modal, for CollA. 
We narrow the discussion to features extracted from our selection of conversation corpora presented in Section~\ref{sec:corpora}, arguing that task-oriented corpora built with particular settings (\textit{e.g.}, interdependency, information asymmetry between participants) that foster collaboration are better suited for identifying CollA features. Both features that have shown significant associations with collaboration quality and those used in individual- and group-level collaboration modeling are included.

We also cover different feature-generation methods (\textit{e.g.}, natural language processing, signal processing, questionnaires) and how these features are exploited (\textit{e.g.}, used directly in modeling or incorporated into high-level construct building) to make the discussion more practical. An overview of features per modality is available in Table \ref{tab:table_features-mod}.

\setlength{\tabcolsep}{2pt}
\begin{table*}[ht]
  \tiny
  \centering
\begin{tabularx}{\textwidth}{|p{1cm}|p{3.4cm}p{3.5cm}|p{2.2cm}p{2.cm}p{3cm}|}
 \cline{2-6}
  \multicolumn{1}{c}{} & \multicolumn{2}{|c|}{\textbf{Group-level Collaboration}} &\multicolumn{3}{c|}{\textbf{Individual Collaborativeness} }
 \\ \cline{2-6}

   \multicolumn{1}{c}{}& \multicolumn{1}{|c}{Group Cohesion and Entrainment} & Teamwork Process and Performance & Engagement & Collaboration Behaviors & Perceived Personality and Role
 \\ \cline{1-6}
  \multirow{1}{1cm}{\textbf{Text \ref{sub:text}}}
  & 
  pronoun usage \citep{enayet2021analyzing}\newline 
  lexical entrainment \citep{rahimi2020entrainment2vec}\newline 
  paralinguistic mimicry \citep{nanninga2017estimating}\newline
  dialogue act Be-Positive \citep{kantharaju2020multimodal}  
  
  & syntactic entrainment embedding \citep {enayet2021analyzing}\newline
  DACTs sequence embedding, sentiment embedding~\citep{enayet2023proactive,enayet2021analyzing}\newline 
  SUBTL score \citep{murray2018predicting}\newline
  lexical cohesion \citep{rahimi2020entrainment2vec}\newline
  dependency parse feature \citep{murray2018predicting}\newline 
  word psycholinguistic score \citep{murray2018predicting} 
  
  & word embedding \citep{li2024multimodal}
  
  &  \textcolor{red}{underexplored}
  
  & BERT embedding \citep{fenech2022perceived}   
    \\ \hline

\multirow{1}{1cm}{\textbf{Audio \ref{sub:audio}}}
  & intensity, frequency, shimmer, jitter \citep{peechatt2024multicollab,litman2016teams}\newline
  turn-taking \citep{sabry2021exploratory,sassier2025toward}\newline
  laughter,backchannels \citep{kantharaju2021social}

  & total overlapping, pause time \citep{hung2010estimating}  
  & audio embedding \citep{li2024multimodal}
  &  \textcolor{red}{underexplored}
  &  total speaking time, pitch, jitter, loudness \citep{peechatt2024multicollab,litman2016teams, sabry2021exploratory}\newline 
  shimmer, harmonics-to-noise \citep{sabry2021exploratory}\newline
  eGeMAPS features \citep{fenech2022perceived}

    \\ \cline{1-6}

\multirow{1}{1cm}{\textbf{Video \ref{sub:video}}} 
  & mutual gaze \citep{kantharaju2021social}\newline
   automatic extracted facial expression, head nods duration \citep{kantharaju2020multimodal}
  &   \textcolor{red}{underexplored}
  &  \textcolor{red}{underexplored}\textcolor{red}{underexplored}
  &  saccade peak velocity \citep{peechatt2024multicollab}
  &  focus of attention without 
mutual  engagement \citep{sabry2021exploratory}\newline 
facial action units from OpenFace \citep{fenech2022perceived}
  
    \\ \hline
    
\multirow{1}{1cm}{\textbf{Sensor \ref{sub:sensor}}}
  & group and individual proxemics and kinesics features \citep{sabry2021exploratory}
  &   bodily motion energy synchrony \citep{kantharaju2021social} 
  &  \textcolor{red}{underexplored}
  &  galvanic skin response (GSR) \citep{peechatt2024multicollab}
  &  traveled distance, kinetic energy, posture expansion, amount of walking, and hand gesture \citep{sabry2021exploratory}
  
    \\ \hline

\multirow{1}{1cm}{\textbf{Cross-modality \ref{sub:cross}}}

   &  representation based leadership \citep{sabry2021exploratory}\newline 
   interpersonal synchrony \citep{sassier2025toward}\newline
   mutual gaze instance during interruption \citep{kantharaju2020multimodal}\newline
   MUMIN coding on social cues \citep{kantharaju2021social}
  &   \textcolor{red}{underexplored}
  &  \textcolor{red}{underexplored}
  &  \textcolor{red}{underexplored}
  &  ratio between successful interruptions and speaking turns \citep{sabry2021exploratory}
  
    \\ \hline

\end{tabularx}
\caption{Features are categorized by modality and their application in group-level and individual-level collaboration analysis, highlighting current trends and potential underexplored areas in CollA research. Detailed discussions of each modality are referenced in their respective sections.}
    \label{tab:table_features-mod}
\end{table*}

\subsection{Text-Based Features}
\label{sub:text}
Text-based features coming from lexical, syntactic, and semantic properties have been studied for CollA. 
The entrainment of \textit{pronoun usage} (\textit{e.g.},  of singular/plural pronoun usage, 1st/2nd/3rd-person pronoun usage) has been applied in team performance level classification \cite{enayet2021analyzing}. 
\textit{Discourse markers} (\textit{e.g.}, ``okay'', ``but'', ``because'') 
signal the communicative function of a phrase (\textit{e.g.}, agreement, disagreement) 
and they can be used as individual-level collaborative behavior features~\cite{koutsombogera2018modeling}, 

Both lexical and syntactic \textit{entrainment}, describing how team members adopt similar speaking styles during conversation, have been studied: syntactic entrainment, calculated using 
automatic part-of-speech tagging,
has been shown to be an effective predictor of team performance, but is expected to be effective only in the late stages of collaboration \cite{enayet2021analyzing}. Lexical entrainment of function words based on LIWC-derived categories of function words \cite{pennebaker2001linguistic} has been used to identify influencers, connectors, and passive members \cite{rahimi2020entrainment2vec} for multiparty collaboration.

BERT-based pretrained models~\cite{devlin2019bert,liu2019roberta,he2020deberta} can be used for text embedding generation in conversation, mapping high-dimensional spaces to low dimensions while retaining only the most effective representations as sparse vectors. This approach has been employed for feature generation in engagement modeling \cite{li2024multimodal} and conflict modeling \cite{enayet2023proactive}.

\subsection{Audio}
\label{sub:audio}
Audio of the collaborator's speech contains many useful features for CollA, such as \textit{intensity}, \textit{frequency}, and their variations (\textit{e.g.}, shimmer, jitter). They can be used to measure speaking energy, pitch, voice quality and excitement, which have been found to be positively correlated with both extreme emotion like frustration \cite{peechatt2024multicollab} and group cohesion \cite{litman2016teams}, and often require cross-modality verification. For frustration identification, \textit{voice features} (\textit{e.g.}, F0, intensity) can be more salient than \textit{visual features}, such as chin raises and brow furrows \cite{peechatt2024multicollab}. Audio features can also automatically be extracted using tools such as OpenSMILE \cite{eyben2010opensmile} and pretrained wav2vec \cite{schneider2019wav2vec}. This has been applied in student engagement prediction~\cite{li2024multimodal}.

\subsection{Video and Sensor Signal } 
\label{sub:video}
\paragraph{Eye Gaze}
Eye gaze refers to the direction and movement of a person's eyes, often used in communication to signal attention, engagement, and turn-taking \cite{kraaij2005ami}.
It helps regulate conversational flow; for example, it has been found that speakers avert their gaze to signal turn initiation and re-establish eye contact to yield the floor \cite{hung2010estimating}. \textit{Mutual gaze}, indicating shared attention, is found to be related to social cohesion \cite{kantharaju2021social}. Additionally, metrics such as \textit{saccade peak velocity}, which can be measured via eye-tracking sensors, have been shown to indicate frustration during collaboration \cite{peechatt2024multicollab}.

\vspace{-0.4cm}

\paragraph{Facial Expression} 
Perceived facial expressions provide insights into participants' emotional states and can contribute to the functional meanings of human interactions. 
For example, \textit{lip corner puller}
, automatically extracted using OpenFace \cite{amos2016openface}, is observed as a more frequent and longer action in high-cohesive segments than in low-cohesive segments for small group meetings, while there are no significant differences for \textit{outer brow raiser} 
and \textit{brow lowerer}
~\cite{kantharaju2020multimodal}. \citet{kantharaju2021social} conduct a comprehensive facial unit study to examine its correlation with high- and low-cohesive segments in the collaboration process.

\paragraph{Head, Hand, Body Motion and Sensor Signals} 
\label{sub:sensor}
Interlocutors synchronize in overall body movement. Bodily motion energy synchrony, a simplified version of interpersonal synchrony, is frequently observed in highly cohesive teams \cite{kantharaju2021social}. \citet{sabry2021exploratory} compute \textit{proxemics} (\textit{e.g.}, interpersonal distance) and \textit{kinesics} (\textit{e.g.}, amount of walking, energy synchrony) features of motion caption to model the emergent leadership and group cohesion. 
Biophysical signals, such as \textit{Galvanic Skin Response} (GSR), provide physiological indicators of states, such as frustration, during collaboration \cite{peechatt2024multicollab}.

\subsection{Dialogic and Cross-Modality}
\label{sub:cross}
Dialogue act (DACT) annotation codes the speaker's intention. \citet{kantharaju2020multimodal} study 15 DACTs following the coding scheme used in the corpus AMI\footnote{\url{https://groups.inf.ed.ac.uk/ami/corpus/Guidelines/dialogue_acts_manual_1.0.pdf}} and find the ``Be-Positive'' DACT as 
highly related to group cohesion. However, as far as we know, no other DACTs have experimentally been proven to be salient in CollA. 

\textit{Turn-taking management} 
can reveal both participant dominance and disengagement \cite{koutsombogera2018modeling}, but requires further validation from other modalities to determine specific functions of turns. The \textit{total pause time} during the collaborative conversation can reflect participants' attentiveness and is consistently high in highly cohesive meetings \cite{hung2010estimating}.

Cross-modality inter-speaker features are intuitively more grounded than unimodal individual features for group analysis. However, for hand-crafted cross-modality features, an early study from  \citet{hung2010estimating} shows that \textit{silent motion} (\textit{i.e.}, visual activity when a person is not talking) is relatively a salient feature in cohesion estimation but does not outperform the total pause time between individual turns, which is an audio-only feature. \textit{Audiovisual synchrony}, either for the same person (\textit{e.g.}, body gestures aligned with speech) or across group members (\textit{e.g.}, interlocutors align with one another in both motion and prosody), can be a good indicator of rapport and comfort \cite{hung2010estimating}.

Both the \textit{number of interruptions} and the \textit{number of mutual gaze instances occurring during interruption} are positively correlated with group cohesion \cite{kantharaju2020multimodal}. \textit{Overlapping speech}, combined with \textit{visual expressions} and \textit{prosodic energy}, helps identify dominance and task cohesion \cite{hung2010estimating}. 
Shared laughter can be observed more frequently and can also last longer in high-cohesion situations \cite{kantharaju2021social}.

\subsection{Main Trends of CollA Features}
Many features have been shown to be statistically significant and are used in collaboration modeling, from collected individual low-level cues to assembled group-level constructs (\textit{e.g.}, entrainment, convergence). 

To better capture group-level collaborative phenomena, some studies explored feature-level fusion across modalities to obtain more grounded features \cite{hung2010estimating,kantharaju2020multimodal,kantharaju2021social}. However, cross-modality features remain underexplored relative to single-modality features in CollA.

We also want to highlight the usage of pretrained models for feature embedding in CollA studies. This approach has been seen a lot in recent CollA, as it enables cross-corpora generalization in the results. However, its effectiveness depends heavily on the robustness and stability of the underlying pretrained models. When applying such models for feature extraction, careful consideration must be given to the alignment between the pretraining data and the intended modeling objectives. For instance, studies have shown that automatically extracted facial action units from pretrained models performed poorly in classifying self-assessed frustration during collaboration \cite{peechatt2024multicollab}, highlighting the importance of validating pretrained feature extractors for specific CollA tasks.

\section{Models}
\label{sec:models}

As discussed in Section \ref{sec:corpora}, existing CollA corpora can include both task-level collaboration annotations as well as segment-level annotations. In this section, we discuss existing collaboration models based on segment-level and task-level annotations, comparing them across their modeling objects, modeling approaches, feature designs, and modality fusion. 

The objective is to understand the possible reasons behind different choices of analysis granularity in the CollA research community and identify valuable future directions. We discuss CollA studies that use at least one of the corpora listed in Section \ref{sec:corpora}. 

\subsection{Predicting Task-Level Annotations}
Task-level collaboration annotations have previously been used to model team collaboration quality \cite{litman2016teams,haberPhotoBookDatasetBuilding2019}, individual collaboration effort \cite{mcduff2017multimodal}, and group cohesion \cite{maman2020game}.

These evaluations are suitable for correlation studies between collaboration and \textbf{task-level}, \textbf{temporal-difference} CollA features, such as entrainment \cite{rahimi2020entrainment2vec, paletz2023speaking}, convergence \cite{rahimi-litman-2018-weighting}, dominance \cite{vogel2023aspects} that change over time during the collaboration process.
We also observe that when modeling with task-level annotations, features are typically aggregated across the entire session (\textit{e.g.}, average, standard deviation, min max value, distribution \cite{walocha2020modeling}). 

Multimodal embeddings of conversations generated by pre-trained models have also been tested in the prediction of task-level collaboration \cite{enayet2021analyzing,enayet2023proactive, rahimi2020entrainment2vec}.
For perceived collaboration quality, scaling data for supervised learning is relatively easier, since task-level evaluations of perceived dimensions can be added to existing task-oriented corpora, enabling model training on combined corpora.
We observe that there are more deep-learning approaches (\textit{e.g.},  LSTM~\cite{enayetImprovingGeneralizabilityCollaborative2023,enayet2023proactive} for conflict prediction, multimodal transformer \cite{fenech2022perceived} for personality modeling) applied to combined, relatively large corpora with task-level evaluation \cite{enayetImprovingGeneralizabilityCollaborative2023,enayet2023proactive}.
We observe that deep-learning approaches (\textit{e.g.}, LSTM~\cite{enayetImprovingGeneralizabilityCollaborative2023,enayet2023proactive} for conflict prediction and multimodal transformers \cite{fenech2022perceived} for personality modeling) are more common in studies using combined, relatively large corpora with task-level evaluations.

\subsection{Predicting Segment-Level Annotations}

Most segment-level annotations are not directly for group-level collaboration quality modeling, but rather for group-level emergent collaboration phenomena modeling \cite{kantharaju2020multimodal,peechatt2024multicollab}. Due to the context-dependent nature of collaboration, existing segment-level direct collaboration annotations can be too scenario-specific to be utilized in further studies of different collaborative situations \cite{richey2016sri}. Collaboration-related aspects with broader applicability to human interaction (\textit{e.g.}, engagement \cite{reverdy2022roomreader}, cohesion \cite{kantharaju2020multimodal}, and frustration \cite{peechatt2024multicollab}) have attracted attention from the research community. 

Segment-level annotations enable research on emergent group behavior and phenomena by leveraging low-level cues and sequential temporal features that capture their context-dependent nature. However, collaboration modeling based on these segment-level annotations still faces several challenges. First, the relatively small, often imbalanced datasets limit the choice of supervised models to classical classifiers such as support vector machines and logistic regression \cite{enayet2021analyzing, kantharaju2021social}.  Second, the automatically extracted multimodal features can have alignment issues, and it is challenging to create a common representation space that preserves cross-modal relationships without losing modality-specific nuances. 

Given data scarcity, solutions have been explored at different levels: at the data level, corpus combination \cite{enayetImprovingGeneralizabilityCollaborative2023,enayet2023proactive},  and synonym replacement in conversational text \cite{enayet2023proactive}; at the feature level, features oversampling \cite{corbellini2023few}.
These methods require a careful adaptation to the modeling object to avoid introducing bias: for example, data augmentation with synonym replacement can delete lexical entrainment in collaboration dynamics. At the model level, graph-based neural network (GNN) methods have achieved state-of-the-art performance on the RoomReader corpus, which contains only 8 hours of recordings \cite{li2024multimodal}, and have remained effective for social interaction classification across several corpora with different task settings \cite{corbellini2023few}. We therefore expect more studies on the potential of GNNs and other graph-learning methods in CollA.

\subsection{State-Of-The-Art Performance}
Direct comparison of results remains challenging due to variations in collaboration tasks, group sizes, contextual settings, analytical approaches, and evaluation metrics.

Model performance in collaboration analysis depends strongly on task formulation and label structure. Model performance in collaboration analysis depends strongly on task formulation and label structure. For example, binary classification tasks, such as high/low group cohesion or high/low arousal, tend to yield higher accuracy, reaching up to 78\% on AMI and PCC \cite{kantharaju2021social}. Imbalanced multiclass problems are more challenging: in four-class group-level Q-code classification, average unweighted accuracy drops to around 49\% on the relatively small SRI corpus \cite{bassiou2016privacy}. Similarly, tasks with well-defined and balanced classes, such as binary engagement detection, can exceed 90\% accuracy with multimodal features \cite{li2024multimodal}.

Overall, deep-learning approaches outperform classical methods in both performance and scalability for multimodal modeling, especially on larger or combined datasets.

For conflict prediction, \citet{rahimi-litman-2018-weighting} achieves 67.74\% accuracy using SVM, while \cite{enayet2021learning} use an LSTM for feature engineering and achieve 73.33\% accuracy on GitHub Issue Dataset\footnote{https://github.com/ayeshaEnayet/DAC-USE}, with training performed on the Teams corpus.

Late fusion strategies, which combine modality-specific embeddings at the decision level, lose modality interactions but have been adopted in recent studies as a trade-off to enable training on several corpora. These results underscore the importance of both task formulation and model architecture in advancing automatic collaboration analysis.

\section{Conclusion and Future Direction}
\label{sec:concl}

This survey provides an overview of recent advances in the analysis of collaboration through human-human conversational corpora. We discuss coding schemes for individual- and group-level collaboration annotation, different task settings for building the CollA corpora, salient multimodal features, and modeling approaches with different granularity. We highlight the evolution from classical statistical methods to deep learning and large language models, as well as the growing integration of multimodality to provide a more nuanced understanding of collaborative processes.

Several directions for future research on collaboration analysis emerge. First, deeper analysis and modeling of individual collaboration strategies using linguistic frameworks 
would enhance our understanding of collaboration dynamics \cite{haberPhotoBookDatasetBuilding2019}. 
Then, more public multimodal corpora for CollA are needed, especially with recording settings that enable group-level multimodal feature capture, such as room-level video \cite{koutsombogera2018modeling} or wearable sociometric badges used in TeamSense corpus\footnote{TeamSense is not a public corpus and is not included in the range of this review.} \cite{zhang2018teamsense}. Zero-few shot and in-context learning approaches have been applied in many recent studies for the CollA of human-machine collaboration systems. These studies often choose an aspect of human collaboration evaluation and aim to align human-machine or machine-only collaborative conversations towards human level, while human conversations are inevitably downgraded in their modality diversity to be comparable. To the best of our knowledge, the use of these methods for CollA within multimodal human conversational data is understudied.

A key insight is that collaboration analysis is inherently task-oriented and interpersonal, which is driving its growing application in human-centered human-machine collaboration systems. Automatic analysis of collaboration through conversational data is a rapidly evolving field. Continued interdisciplinary efforts, combining linguistics, computer science, psychology, and education, will be essential to address challenges and unlock the full potential of collaborative technologies in both research and real-world applications.

\section{Acknowledgements}
Comments from the reviewers are greatly appreciated. Special thanks also go to Aina Garí Soler and Elodie Etienne for their comments. This work was  funded by the 
ANR-23-CE23-0033-01 SINNet project and additional support was provided by the ANR under the France 2030 program PRAIRIE (ANR-23-IACL-0008).
\section{Bibliographical References}\label{sec:reference}

\bibliographystyle{lrec2026-natbib}
\bibliography{lrec}

\begin{thebibliography}{13}
\expandafter\ifx\csname natexlab\endcsname\relax\def\natexlab#1{#1}\fi

\bibitem[{Braley and Murray(2018)}]{braleyGroupAffectPerformance2018}
Braley, McKenzie and Murray, Gabriel. 2018.
\newblock \href {https://doi.org/10.1145/3279981.3279985} {\emph{{The Group Affect and Performance (GAP) Corpus}}}.
\newblock Association for Computing Machinery, GIFT'18.

\bibitem[{Cavicchio and Poesio(2012)}]{cavicchio2012rovereto}
Cavicchio, Federica and Poesio, Massimo. 2012.
\newblock \href {https://www.jstor.org/stable/pdf/41486068.pdf?casa_token=QMxsCZgZFdQAAAAA:ShjO5-bjI5rv-_o20d5fQoJm4jR-eZCjhllnMOISVuKif0Lbm4wLUYk8s5zlpV85rneVQGxqZq8Trrnu_ZkbFCkOHraHcUjy28spsi4XJj6ZjGXfiEHryw} {\emph{{The Rovereto Emotion and Cooperation Corpus: A New Resource to Investigate Cooperation and Emotions}}}.
\newblock Springer.

\bibitem[{Haber et~al.(2019)Haber, Baumg{\"a}rtner, Takmaz, Gelderloos, Bruni, and Fern{\'a}ndez}]{haberPhotoBookDatasetBuilding2019}
Haber, Janosch and Baumg{\"a}rtner, Tim and Takmaz, Ece and Gelderloos, Lieke and Bruni, Elia and Fern{\'a}ndez, Raquel. 2019.
\newblock \href {https://doi.org/10.18653/v1/P19-1184} {\emph{The PhotoBook Dataset: Building Common Ground through Visually-Grounded Dialogue}}.
\newblock Association for Computational Linguistics.

\bibitem[{Kantharaju and Pelachaud(2021)}]{kantharaju2021social}
Kantharaju, Reshmashree B and Pelachaud, Catherine. 2021.
\newblock \href {https://dl.acm.org/doi/pdf/10.1145/3472306.3478362?} {\emph{{Social Signals of Cohesion in Multi-Party Interactions}}}.

\bibitem[{Koutsombogera and Vogel(2018)}]{koutsombogera2018modeling}
Koutsombogera, Maria and Vogel, Carl. 2018.
\newblock \href {https://aclanthology.org/L18-1466.pdf} {\emph{{Modeling Collaborative Multimodal Behavior in Group Dialogues: The MULTISIMO Corpus}}}.

\bibitem[{Kraaij et~al.(2005)Kraaij, Hain, Lincoln, and Post}]{kraaij2005ami}
Kraaij, Wessel and Hain, Thomas and Lincoln, Mike and Post, Wilfried. 2005.
\newblock \href {https://d1wqtxts1xzle7.cloudfront.net/50793769/The_AMI_meeting_corpus20161208-17868-1xaka8f-libre.pdf?1481255943=&response-content-disposition=inline%3B+filename%3DThe_AMI_Meeting_Corpus.pdf&Expires=1761252899&Signature=JjbD2XbLv49WQEKYayDBIODGWUW-A~iXfbpvO1teXCYfSvq9fxdyOBJIOJabi05OQo9ldR~1Z2Brgn7YgV4LjkLzPIIia-pCZ6GhhehTIEj4VsWPfq0-mphCw~KWHgp64rLmMgonH49Za050nTiX9AoJxZ~4qqErULzet-EOcmdqiyYGBUAJQAWtOMVUG~fgBSsbJ7kGRHkPO663R8erj5fLwdj2WGtM0ljaPuDtdRKgBWzYs6ikM1zORRlZQeFDJckgzJ8VluTdUVD8DsGMJRDMfD7GMjKPUbY7wYjxouenPRnLhPjEUHqfEabdTmEZKWBIUCTYcUQQdZXGRiPufA__&Key-Pair-Id=APKAJLOHF5GGSLRBV4ZA} {\emph{{The AMI Meeting Corpus}}}.

\bibitem[{Litman et~al.(2016)Litman, Paletz, Rahimi, Allegretti, and Rice}]{litman2016teams}
Litman, Diane and Paletz, Susannah and Rahimi, Zahra and Allegretti, Stefani and Rice, Caitlin. 2016.
\newblock \href {https://aclanthology.org/D16-1149.pdf} {\emph{{The Teams Corpus and Entrainment in Multi-party Spoken Dialogues}}}.

\bibitem[{Maman et~al.(2020)Maman, Ceccaldi, Lehmann-Willenbrock, Likforman-Sulem, Chetouani, Volpe, and Varni}]{maman2020game}
Maman, Lucien and Ceccaldi, Eleonora and Lehmann-Willenbrock, Nale and Likforman-Sulem, Laurence and Chetouani, Mohamed and Volpe, Gualtiero and Varni, Giovanna. 2020.
\newblock \href {https://ieeexplore.ieee.org/stamp/stamp.jsp?arnumber=9127943} {\emph{{Game-on: A Multimodal Dataset for Cohesion and Group Analysis}}}.
\newblock IEEE.

\bibitem[{McDuff et~al.(2017)McDuff, Thomas, Czerwinski, and Craswell}]{mcduff2017multimodal}
McDuff, Daniel and Thomas, Paul and Czerwinski, Mary and Craswell, Nick. 2017.
\newblock \href {https://dl.acm.org/doi/pdf/10.1145/3136755.3136813?} {\emph{{Multimodal Analysis of Vocal Collaborative Search: A Public Corpus and Results}}}.

\bibitem[{Olshefski et~al.(2020)Olshefski, Lugini, Singh, Litman, and Godley}]{olshefski2020discussion}
Olshefski, Christopher and Lugini, Luca and Singh, Ravneet and Litman, Diane and Godley, Amanda. 2020.
\newblock \href {https://aclanthology.org/anthology-files/pdf/lrec/2020.lrec-1.130.pdf} {\emph{{The Discussion Tracker Corpus of Collaborative Argumentation}}}.

\bibitem[{Peechatt et~al.(2024)Peechatt, Alm, and Bailey}]{peechatt2024multicollab}
Peechatt, Michael and Alm, Cecilia Ovesdotter and Bailey, Reynold. 2024.
\newblock \href {https://aclanthology.org/2024.lrec-main.1023.pdf} {\emph{{ MULTICOLLAB: A Multimodal Corpus of Dialogues for Analyzing Collaboration and Frustration in Language}}}.

\bibitem[{Reverdy et~al.(2022)Reverdy, Russell, Duquenne, Garaialde, Cowan, and Harte}]{reverdy2022roomreader}
Reverdy, Justine and Russell, Sam O’Connor and Duquenne, Louise and Garaialde, Diego and Cowan, Benjamin R and Harte, Naomi. 2022.
\newblock \href {https://aclanthology.org/2022.lrec-1.268.pdf} {\emph{{Roomreader: A Multimodal Corpus of Online Multiparty Conversational Interactions}}}.

\bibitem[{Richey et~al.(2016)Richey, D'Angelo, Alozie, Bratt, and Shriberg}]{richey2016sri}
Richey, Colleen and D'Angelo, Cynthia M and Alozie, Nonye and Bratt, Harry and Shriberg, Elizabeth. 2016.
\newblock \href {https://www.isca-archive.org/interspeech_2016/richey16_interspeech.pdf} {\emph{{The SRI Speech-Based Collaborative Learning Corpus.}}}
\newblock ISLRN \href{https://www.islrn.org/resources/199-041-455-836-2}{199-041-455-836-2}.

\end{thebibliography}

\section{Language Resource References}
\label{lr:ref}
\bibliographystylelanguageresource{lrec2026-natbib}
\bibliographylanguageresource{lreclang}
\section*{Appendix A - Corpora Selection Process for CollA: A Relatively Less-Resourced Domain}
\begin{figure}[H]
    \centering
    \includegraphics[width=\linewidth]{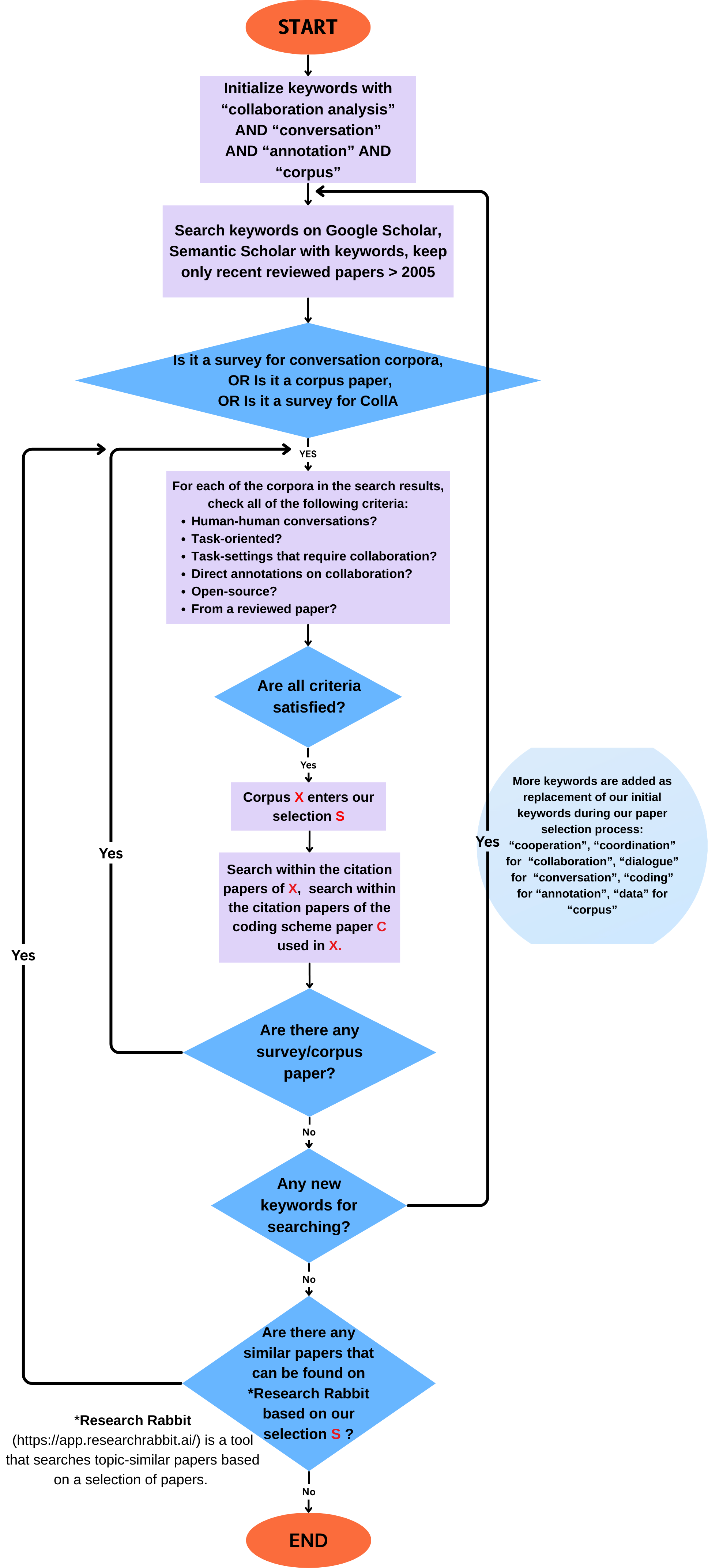}
    \caption{Human-human conversation corpus building with manual annotations is a costly process. It was particularly difficult to find corpora with direct annotations on collaboration. Our recursive approach is time-consuming, but the boundary is clearly defined, which results in a relatively grounded and complete selection for an overview of CollA corpora.}
    \label{fig:data_select}
\end{figure}

\end{document}